# Verdict Accuracy of Quick Reduct Algorithm using Clustering, Classification Techniques for Gene Expression Data

T.Chandrasekhar[1], K.Thangavel[2] and E.N.Sathishkumar[3]

[1] Department of Computer Science, Periyar University,
Salem, Tamilnadu-636 011, India
*ch_ansekh80@rediffmail.com*

[2] Department of Computer Science, Periyar University,
Salem, Tamilnadu-636 011, India
*drktvelu@yahoo.com*

[3] Department of Computer Science, Periyar University,
Salem, Tamilnadu-636 011, India
*en.sathishkumar@yahoo.in*

**Abstract**

In most gene expression data, the number of training samples is very small compared to the large number of genes involved in the experiments. However, among the large amount of genes, only a small fraction is effective for performing a certain task. Furthermore, a small subset of genes is desirable in developing gene expression based diagnostic tools for delivering reliable and understandable results. With the gene selection results, the cost of biological experiment and decision can be greatly reduced by analyzing only the marker genes. An important application of gene expression data in functional genomics is to classify samples according to their gene expression profiles. Feature selection (FS) is a process which attempts to select more informative features. It is one of the important steps in knowledge discovery. Conventional supervised FS methods evaluate various feature subsets using an evaluation function or metric to se       lect only those features which are related to the decision classes of the data under consideration. This paper studies a feature selection method based on rough set theory. Further K-Means, Fuzzy C-Means (FCM) algorithm have implemented for the reduced feature set without considering class labels. Then the obtained results are compared with the original class labels. Back Propagation Network (BPN) has also been used for classification. Then the performance of K-Means, FCM, and BPN are analyzed through the confusion matrix. It is found that the BPN is performing well comparatively.

## 1. Introduction

Feature selection is the process of choosing the most appropriate features when creating the model of the process. Most of the feature selection methods are applied across the entire data set. Once such genes are chosen, the creation of classifiers on the basis of the genes is another undertaking. If we survey the established investigations in this field, we will find that almost all the accurate classification results are obtained based on more than two genes. Rough sets have been used as a feature selection methods by many researchers among them Jensen and Schen, Zhong et al, Wang and Hu et al. The Rough set approach to feature selection consists in selecting a subset of features which can predict the classes as well as the original set of features. The optimal criterion for Rough set feature selection is to find shortest or minimal reducts while obtaining high quality classifiers based on the selected features. Here we propose a feature selection method based on rough set theory for reducing genes from large gene expression database [1, 4].

Discriminant analysis is now widely used in bioinformatics, such as distinguishing cancer tissues from normal tissues. A problem with gene expression analysis or with any large dimensional data set is often the selection of significant variables (feature selection) within the data set that would enable accurate classification of the data to some output classes. These variables may be potential diagnostic markers too. There are good reasons for reducing the large number of variables:

1) An opportunity to scrutinize individual genes for further medical treatment and drug development. 2) Dimension reduction to reduce the computational cost. 3) Reducing the number of redundant and unnecessary variables can improve inference and classification. 4) More interpretable features or characteristics that can help identify and monitor the target diseases or function types [5].

The rest of the paper is organized as follows: Section 2, briefs about the Rough set theory. Section 3 describes the clustering techniques. Section 4 briefs about classification techniques. Section 5 explains briefly about experimental analysis and results. Section 6 presents a conclusion for this paper.

## 2. Rough Set Theory

Rough set theory (Pawlak, 1991) is a formal mathematical tool that can be applied to reducing the dimensionality of datasets. The rough set attribute reduction method removes redundant input attributes from datasets of discrete values, all the while making sure that no information is lost. The approach is fast and efficient, making use of standard operations from conventional set theory [3].

**Definition:** Let $U$ be a universe of discourse, $X \subseteq U$, and $R$ is an equivalence relation on $U$. $U/R$ represents the set of the equivalence class of $U$ induced by $R$. The *positive region* of $X$ on $R$ in $U$, is defined as $pos(R,X) = U \{Y \in U/R \mid Y \subseteq X\}$.

The partition of $U$, generated by $IND$ $(P)$ is denoted $U/P$. If $(x, y) \in IND$ $(P)$, then $x$ and $y$ are indiscernible by attributes from $P$. The equivalence classes of the P-indiscernibility relation are denoted $[x]p$. The indiscernibility relation is the mathematical basis of rough set theory. Let $X \subseteq U$, the P-lower approximation $\underline{P}X$ and P-upper approximation $\overline{P}X$ of set $X$ can be defined as:

$$\underline{P}X = \{ x \in U \mid [x]p \subseteq X \} \quad (1)$$

$$\overline{P}X = \{ x \in U \mid [x]p \cap X \neq \varphi \} \quad (2)$$

Let $P, Q \subseteq A$ be equivalence relations over $U$, then the positive, negative and boundary regions can be defined as:

$$POS_P(Q) = \bigcup_{X \in U/Q} \underline{P}X \quad (3)$$

$$NEG_P(Q) = U - \bigcup_{X \in U/Q} \overline{P}X \quad (4)$$

$$BND_P(Q) = \bigcup_{X \in U/Q} \overline{P}X - \bigcup_{X \in U/Q} \underline{P}X \quad (5)$$

An important issue in data analysis is discovering dependencies between attributes dependency can be defined in the following way. For $P, Q \subseteq A$, $P$ depends totally on $Q$, if and only if $IND$ $(P) \subseteq IND$ $(Q)$. That means that the partition generated by $P$ is finer than the partition generated by $Q$. We say that $Q$ depends on $P$ in a degree $0 \leq k \leq 1$ denoted $P \Rightarrow_k Q$, if

$$k = \gamma_P(Q) = \frac{|POS_P(Q)|}{|U|} \quad (6)$$

If $k = 1$, $Q$ depends totally on $P$, if $0 \leq k \leq 1$, $Q$ depends partially on $P$, and if $k=0$ then $Q$ does not depend on $P$. In other words, $Q$ depends totally (partially) on $P$, if all (some) objects of the universe $U$ can be certainly classified to blocks of the partition $U/Q$, employing $P$. In a decision system the attribute set contains the condition attribute set $C$ and decision attribute set $D$, i.e. $A = C \cup D$. The degree of dependency between condition and decision attributes, $\gamma c(D)$, is called the quality of approximation of classification, induced by the set of decision attributes[6,10].

### 2.1 Quick Reduct Algorithm

The reduction of attributes is achieved by comparing equivalence relations generated by sets of attributes. Attributes are removed so that the reduced set provides the same quality of classification as the original. A reduct is defined as a subset R of the conditional attribute set C such that $\gamma R(D) = \gamma C(D)$. A given dataset may have many attribute reduct sets, so the setR of all reducts is defined as:

$$Rall = \{X \mid X \subseteq C, \gamma X(D) = \gamma C(D);$$
$$\gamma X - \{a\}(D) \neq \gamma X(D), \forall a \in X\}. \quad (7)$$

The intersection of all the sets in Rall is called the core, the elements of which are those attributes that cannot be eliminated without introducing more contradictions to the representation of the dataset. For many tasks (for example, feature selection), a reduct of minimal cardinality is ideally searched for. That is, an attempt is to be made to locate a single element of the reduct set $Rmin \subseteq Rall$:

$$Rmin = \{X \mid X \in Rall, \forall Y \in Rall, |X| \leq |Y|\}. \quad (8)$$

The Quick Reduct algorithm shown below[8, 9], it searches for a minimal subset without exhaustively generating all possible subsets. The search begins with an empty subset; attributes which result in the greatest increase in the rough set dependency value are added iteratively. This process continues until the search produces its maximum possible dependency value for that dataset ($\gamma c(D)$). Note that this type of search does not guarantee a minimal subset and may only discover a local minimum.

QUICKREDUCT(C, D)
C, the set of all conditional features;
D, the set of decision features.
(a) R ← { }
(b) **Do**
(c) T ← R
(d) $\forall x \in (C-R)$
(e) **if** $\gamma R \cup \{x\} (D) > \gamma T (D)$

Where γR(D)=card(POSR(D)) / card(U)
(f)　　　T ← R∪{x}
(g) R ← T
(h) **until**γR(D) = = γC(D)
(i) **return** R

It starts off with an empty set and adds in turn, one at a time, those attributes that result in the greatest increase in the rough set dependency metric, until this produces its maximum possible value for the dataset. Other such techniques may be found in [8, 9]

## 3. Clustering Techniques

Clustering is the process of grouping data into clusters, where objects within each cluster have high similarity, but are dissimilar to the objects in other clusters. Similarities are assessed based on the attributes values that best describes the objects. Often distance measures are used for the purpose. Clustering has its roots in many areas, including data mining, statistics, biology, and machine learning. In this work K-Means, FCM and BPN algorithms which are used to classify the data.

3.1 K-Means Algorithm

*K-M*eans algorithm (MacQueen, 1967) is one of a group of algorithms called partitioning methods. The k-mean algorithm is very simple and can be easily implemented in solving many practical problems. The k-means algorithm is the best-known squared error-based clustering algorithm [11].

Consider the data set with 'n' objects,
　　　i.e., S = {$x_i$ : 1 ≤ i ≤ n}.

1) Initialize a k-partition randomly or based on some prior knowledge.
　　　i.e. {$C_1$, $C_2$, $C_3$,……., $C_k$ }.
2) Calculate the cluster prototype matrix M (distance matrix of distances between k-clusters and data objects).
M = {$m_1, m_2, m_3, …., m_k$ }
Where $m_i$ is a column matrix 1× n .
3) Assign each object in the data set to the nearest cluster - $C_m$
i.e. $x_j ∈ C_m$ if ‖ $x_j$ - $C_m$ ‖ ≤ ‖ $x_j$ – $C_i$ ‖ ∀ 1 ≤ j ≤ k , j ≠m
　　　　　Where j=1, 2, 3, ……., n.
4) Calculate the average of each cluster and change the k-cluster centers by their averages.
5) Again calculate the cluster prototype matrix M.
6) Repeat steps 3, 4 and 5 until there is no change for each cluster.

The k-means algorithm is the most extensively studied clustering algorithm and is generally effective in producing good results. The major drawback of this algorithm is that it produces different clusters for different sets of values of the initial centroids. Quality of the final clusters heavily depends on the selection of the initial centroids [12].

3.2 Fuzzy C Means

Fuzzy clustering allows each feature vector to belong to more than one cluster with different membership degrees (between 0 and 1) and vague or fuzzy boundaries between clusters. Fuzzy c-means (FCM) is a method of clustering which allows one piece of data to belong to two or more clusters. This method (developed by Dunn in 1973 and improved by Bezdek in 1981) is frequently used in pattern recognition [15].

*Algorithm Steps:*

Step-1: Randomly initialize the membership matrix using this equation,

$$\sum_{j=1}^{C} \mu_j(x_i) = 1 \quad i = 1,2,…..k \quad (9)$$

Step-2: Calculate the Centroid using equation,

$$Cj = \frac{\sum_i [\mu_j(x_i)]^m x_i}{\sum_i [\mu_j(x_i)]^m}$$
(10)

Step-3: Calculate dissimilarly between the data points and Centroid using the Euclidean distance.
Step-4: Update the New membership matrix using the equation,

$$\mu_j(x_i) = \frac{[\frac{1}{d_{ji}}]^{1/m-1}}{\sum_{k=1}^{c}[\frac{1}{d_{ki}}]^{1/m-1}}$$

(11)
Here **m** is a fuzzification parameter, The range **m** is always {1.25, 2}
Step-5: Go back to Step 2, unless the centroids are not changing.

## 4. Classification Techniques

4.1 Back Propagation Networks (BPN)

BPN is an information-processing paradigm that is inspired by the way biological nervous systems[13,14],

such as the brain, process information. The key element of this paradigm is the novel structure of the information processing system. It is composed of a large number of highly interconnected processing elements (neurons) working in unison to solve specific problems.

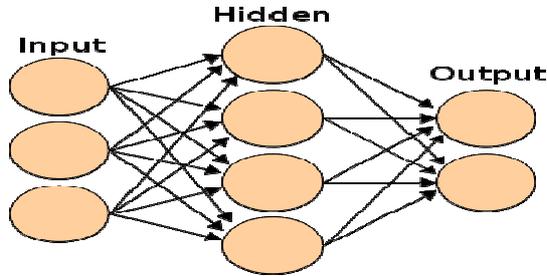

Fig 1: BPN Architecture

Developing a neural network involves first training the network to carry out the desired computations. The feed-forward neural network architecture is commonly used for supervised learning. Feed-forward neural networks contain a set of layered nodes and weighted connections between nodes in adjacent layers. Feed-forward networks are often trained using a back propagation-learning scheme. Back propagation learning works by making modifications in weight values starting at the output layer then moving backward through the hidden layers of the network. Neural networks have been criticized for their poor interpretability, since it is difficult for humans to interpret the symbolic meaning behind the learned weights. Advantages of neural networks, however, include their high tolerance to noisy data as their ability to classify patternson which they have not been trained [13,14].

## 5. Experimental Results

### 5.1 Data Sets

We use four datasets: leukemia, breast cancer, lung cancer and prostate cancer which are available in the website: http://datam.i2r.a-star.edu.sg/datasets/krbd/, [2]. the gene number and class contained in four datasets are listed in Table 1.

Table1: Summary of the four gene expression datasets.

| Dataset | #Gene | Class |
|---|---|---|
| Leukemia | 7129 | ALL/AML |
| Prostate Cancer | 12600 | Tumor/Normal |
| Breast Cancer | 24481 | Relapse/Non Relapse |
| Lung Cancer | 7129 | Tumor/Normal |

The data studied by rough sets are mainly organized in the form of decision tables. One decision table can be represented as $S = (U, A=C \cup D)$, where $U$ is the set of samples, $C$ the condition attribute set and $D$ the decision attribute set. We can represent every gene expression data with the decision table like Table 2.

Table2. Microarray data decision table.

| Samples | Condition attributes(genes) | | | | Decision attributes |
|---|---|---|---|---|---|
| | Gene 1 | Gene 2 | ... | Gene n | Class label |
| 1 | g(1,1) | g(1,2) | … | g(1,n) | Class(1) |
| 2 | g(2,1) | g(2,2) | … | g(2,n) | Class(2) |
| … | … | … | … | … | … |
| m | g(m,1) | g(m,2) | … | g(m,n) | Class(m) |

In the decision table, there are $m$ samples and $n$ genes. Every sample is assigned to one class label. Each gene is a condition attribute and each class is a decision attribute. $g(x, y)$ signifies the expression level of gene $y$ in sample $x$. [2].

### 5.2 Data Pre-processing, Gene Selection

Before applying feature selection algorithm all the conditional attributes (samples) are discretized using K-Means discretization [16]. Let us considered $U$ is the set of samples, $C$ the condition attribute set and $D$ the decision attribute set. By applying Quick Reduct Algorithm, In prostate gene dataset, gene #20 and #11154 are identified, where as in leukemia dataset gene #4 and #3252 are identified, in breast cancer dataset gene #3 and #22019 are identified, finally in lung cancer dataset gene #4817 as best attribute for finding appropriate decision.

Table 3: Features selected by Quick Reduct Algorithm

| Gene Data | Identified Attributes (Genes) |
|---|---|
| Leukemia Cancer | #4, #3252 |
| Prostate Cancer | #20, #11154 |
| Breast Cancer | #3, #22019 |
| Lung Cancer | #4817 |

### 5.3 Classification Performance

In this section the selected data is clustered by the K-Means and FCM algorithm. The data presented in Table 4 and 5 shows the classification performance of True Positive (TP) rate, True Negative (TN) rate, False Positive (FP) rate, and False Negative (FN) rate as previously described. Table 5 shows classification performance of Back Propagation Network. Results are presented both in terms of classification accuracy and classification error [7].

Table 4: K-Means Classification Performance Rate

| Gene Data | K-Means | | | |
|---|---|---|---|---|
| | TP | FP | TN | FN |
| Leukemia Cancer | 1.0000 | 0.0000 | 0.1428 | 0.8571 |
| Prostate Cancer | 0.3750 | 0.6250 | 0.0769 | 0.9230 |
| Breast Cancer | 0.6667 | 0.3333 | 0.4286 | 0.5714 |
| Lung Cancer | 0.9883 | 0.0116 | 0.0000 | 1.0000 |

Table 5: FCMs Classification Performance Rate

| Gene Data | FCM | | | |
|---|---|---|---|---|
| | TP | FP | TN | FN |
| Leukemia Cancer | 1.0000 | 0.0000 | 0.1428 | 0.8571 |
| Prostate Cancer | 0.1250 | 0.8750 | 0.0769 | 0.9230 |
| Breast Cancer | 0.6667 | 0.3333 | 0.4285 | 0.5714 |
| Lung Cancer | 0.9883 | 0.0116 | 0.0000 | 1.0000 |

When comparing classification results, where the BPN method shows a high in classification accuracy, which is demonstrated in Fig. 2.

Table 6: K-Means, FCM and BPN Classification Accuracy

| Gene Data | K-Means | FCM | BPN |
|---|---|---|---|
| Leukemia Cancer | 0.9412 | 0.9412 | 1.0000 |
| Prostate Cancer | 0.7143 | 0.6190 | 0.8000 |
| Breast Cancer | 0.6315 | 0.6315 | 0.9167 |
| Lung Cancer | 0.9896 | 0.9896 | 1.0000 |

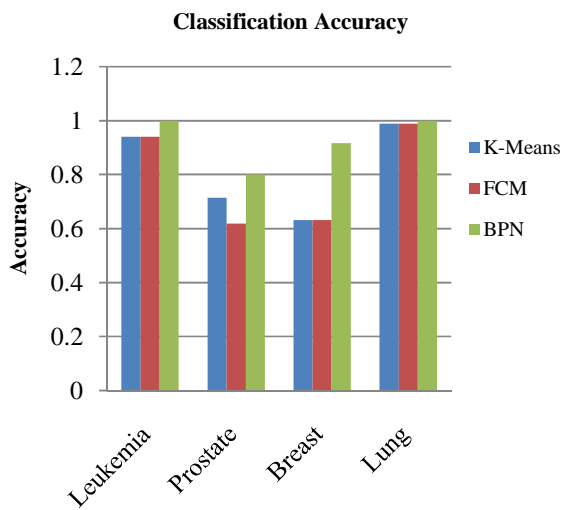

Fig 2: K-Means, FCM and BPN Classification Accuracy

Table 7: K-Means, FCM and BPN Classification Error

| Gene Data | K-Means | FCM | BPN |
|---|---|---|---|
| Leukemia Cancer | 0.0588 | 0.0588 | 0.0000 |
| Prostate Cancer | 0.2857 | 0.3809 | 0.2000 |
| Breast Cancer | 0.3684 | 0.3684 | 0.1667 |
| Lung Cancer | 0.0104 | 0.0104 | 0.0000 |

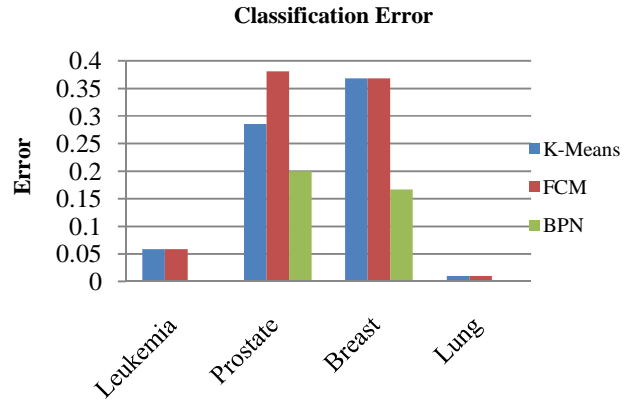

Fig 3: K-Means, FCM and BPN Classification Error

Fig 2 and 3 demonstrated the classification accuracy and error rate of Quick reduct algorithm.

## 6. Conclusion

In this paper, Quick reduct algorithm based on rough set theory has been studied for gene expression datasets. The reduced feature set has been used to cluster the data using K-Means and FCM algorithms with considering decision attributes. The performance was evaluated using confusion matrix with positive and negative class values. Further, the selected features with class labels were classified using Back Propagation Network. It was observed that the performance of the BPN is significant.